\pdfoutput=1

\documentclass[11pt]{article}
\usepackage{CJKutf8}
\usepackage{acl}
\usepackage{authblk}

\usepackage{times}
\usepackage{latexsym}
\usepackage[linesnumbered,ruled,vlined]{algorithm2e}

\usepackage[T1]{fontenc}

\usepackage[utf8]{inputenc}

\usepackage{microtype}

\usepackage{inconsolata}

\usepackage{graphicx}
\usepackage{float}
\usepackage{multirow}

\title{MagicGeo: Training-Free Text-Guided Geometric Diagram Generation}

\begin{document}

\author{
 Junxiao Wang$^1$\textsuperscript{*}
 \quad Ting Zhang$^1$\textsuperscript{*}
 \quad Heng Yu$^1$
 \quad Jingdong Wang$^2$
 \quad Hua Huang$^1$\textsuperscript{†} \\
 {$^1$Beijing Normal University \quad $^2$Baidu}\\
 {}
}


\twocolumn[{
\renewcommand\twocolumn[1][]{#1}
\maketitle
\vspace{-0.45cm}
\centering
\includegraphics[width=0.99\textwidth]{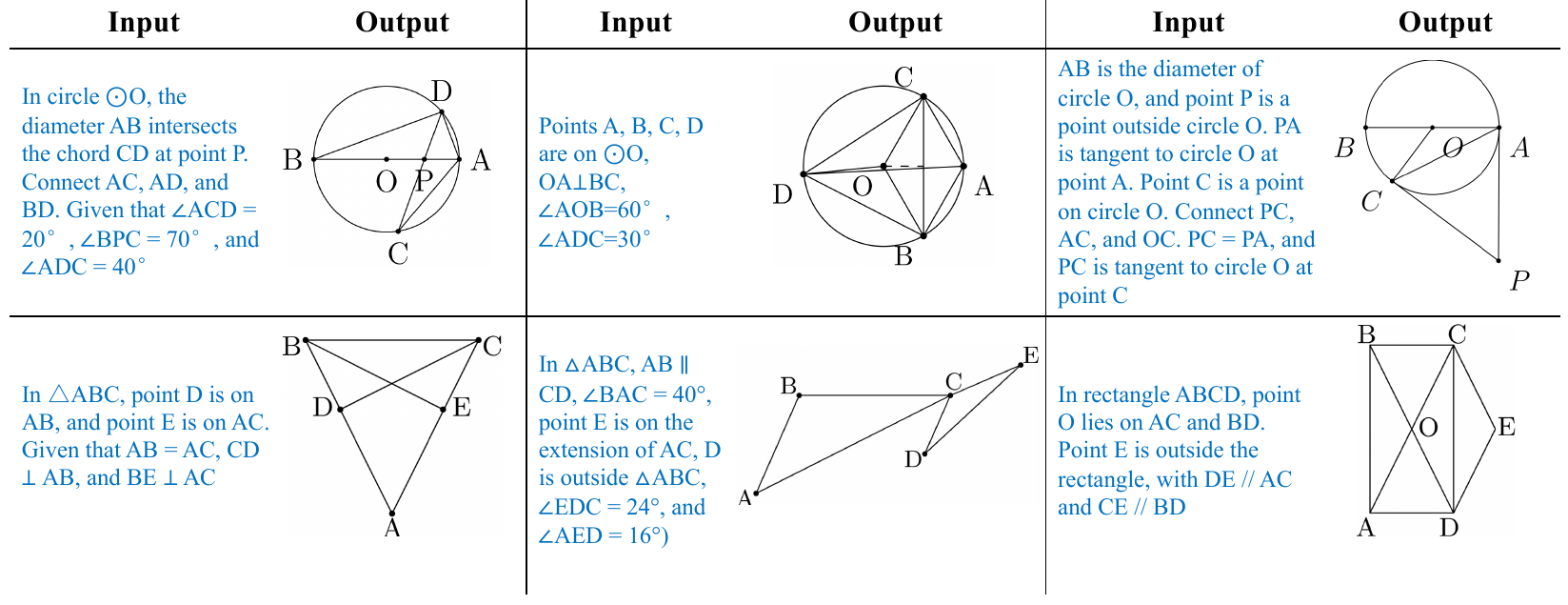}
\captionsetup{type=figure}
\caption{\emph{MagicGeo} has the capability to generate accurate complex geometric diagrams from natural language. }
\vspace{-0.4cm}
\label{fig:teaser}
\vspace{1cm}
}]



{
  \renewcommand{\thefootnote}%
    {\fnsymbol{footnote}}
  \footnotetext[1]{Equal contribution.}
  \footnotetext[2]{Corresponding author.}
}

\begin{abstract}
Geometric diagrams are critical in conveying mathematical and scientific concepts, yet traditional diagram generation methods are often manual and resource-intensive. While text-to-image generation has made strides in photorealistic imagery, creating accurate geometric diagrams remains a challenge due to the need for precise spatial relationships and the scarcity of geometry-specific datasets. This paper presents MagicGeo, a training-free framework for generating geometric diagrams from textual descriptions. MagicGeo formulates the diagram generation process as a coordinate optimization problem, ensuring geometric correctness through a formal language solver, and then employs coordinate-aware generation.
The framework leverages the strong language translation capability of large language models, while formal mathematical solving ensures geometric correctness. 
We further introduce MagicGeoBench, a benchmark dataset of 220 geometric diagram descriptions, and demonstrate that MagicGeo outperforms current methods in both qualitative and quantitative evaluations. This work provides a scalable, accurate solution for automated diagram generation, with significant implications for educational and academic applications.
\end{abstract}

\section{Introduction}

"A picture is worth a thousand words" is a widely recognized proverb in literature. 
Specifically, diagram, as a form of picture, is essential in conveying information and have long been utilized across fields such as science and engineering. Extensive research~\cite{larkin1987diagram,stenning1995cognitive} demonstrates that diagrams often outperform text in solving determinate problems. Prominent figures like Einstein and Hadamard have famously asserted that they do not "think in words"~\cite{larkin1987diagram}. Furthermore,~\citet{stenning1995cognitive} argues that text permits expression of
ambiguity in the way that diagrams cannot easily accommodate. This paper focuses on the task of converting descriptions into structured diagrams, with particular emphasis on geometric diagrams, which play a critical role in mathematics and science. This task serves as a foundational step toward advancing diagram generation for scientific textbooks.

Traditional geometric diagram construction is closely associated with a suite of graphic drawing tools, such as Cinderella~\cite{yu2015automatic}, Geometry Expert~\cite{chou1996introduction}, Z+Z Super Sketchpad~\cite{zhang2007free}, and WinGCLC~\cite{janivcic2003wingclc,szirmay2003proceedings}. These tools offer interactive platforms for drawing geometric figures. However, they are burdened by the need for manual input, which is both time-consuming and resource-intensive.
This paper presents the development of an automatic, text-guided geometric diagram generation system, eliminating the manual effort typically involved. Such a system holds significant potential for streamlining diagram creation, offering considerable utility in the preparation of educational resources.

Recent advancements in text-to-image generation have achieved notable progress in synthesizing photorealistic images~\cite{cao2024survey, zhou2023vision+}. However, these methods, trained on large datasets of natural image-text pairs, often struggle with diagram generation. Efforts to address this challenge include DiagrammerGPT~\cite{zala2023diagrammergpt}, which proposes a two-stage framework using layout as an intermediary to enable spatial control, and AutomaTikZ~\cite{belouadi2023automatikz}, which leverages the TikZ graphic language to autonomously generate scientific figures from captions. Despite these advances, both approaches rely on supervised training data, which limits their generalizability. Furthermore, the scarcity of geometry-specific image-text pairs relative to general image-text corpora makes it difficult to learn the semantic and structural logic of geometric layouts directly from natural language inputs.

In this paper, we introduce MagicGeo, a framework for the automatic generation of text-to-geometric diagrams in a training-free manner, thereby sidestepping the need for paired geometry-text datasets. We focus on geometric diagram as it stands out due to its stringent precision requirement, that is, properties such as parallelism, orthogonality, and degree constraint must be rigorously maintained. Given that even minor inaccuracies are immediately noticeable, this task poses significant challenges within image generation.

Our key insight is that correctness hinges on the precise placement of points. Once the point locations are accurate, constructing the geometry becomes straightforward, such as connecting points with lines or drawing circles. Drawing inspiration from computational geometry methods used in geometry theorem provers~\cite{wu2008decision}, we model diagram generation as a set of polynomial equations based on point coordinates.

While large language models (LLMs) exhibit impressive capabilities in language understanding and reasoning, they are not inherently equipped to solve complex multi-constraint tasks~\cite{kambhampati2024llms}. As a result, directly using LLMs to solve for point coordinates leads to errors and hallucinations.
Instead, we turn to leverage LLM's strengths in translation to convert geometry texts into key formal information. This information is then used to formulate an optimization problem, which is solved algorithmically to ensure that the geometric constraints are satisfied.

To this end, MagicGeo operates in three distinct stages: 1) Autoformalization with LLM: LLMs interpret the geometry description and translate it into an optimization problem, defining a set of constraints with respect to the point coordinates. 2) Solving with Verification: Computational geometry principles are applied to search for one solution that satisfies all constraints; if no solution is found, the system reverts to the autoformalization step to re-extract the necessary information. 3) Coordinate-aware generation: We employ point coordinates to generate TikZ language, which serves as an intermediary representation for the creation of the corresponding geometric diagram.

To advance the evaluation of text-to-geometric diagram generation and promote further research, we present MagicGeoBench, a real-world dataset containing 220 plane geometry descriptions sourced from middle school math exams. 
Empirical results demonstrate that MagicGeo significantly outperforms state-of-the-art baselines, both qualitatively and quantitatively.
Figure~\ref{fig:teaser} illustrates several examples.
We also explore its potential for diagram editing, showcasing how the diagrams can be tailored to user preferences, thereby enhancing practical utility.
While our experiments focus on plane geometry,
the underlying methodology is highly extensible to other geometric branches, such as analytical and solid geometry.
Our current goal is to demonstrate the efficacy of the propose concept, which 
we believe will foster broader exploration and inspire further innovation in the field.

In summary, our key contributions are:
\begin{itemize}

\item We propose a novel perspective that frames geometric diagram generation as a well-defined optimization problem, enhancing its tractability within the zero-shot capabilities of LLMs.

\item We present MagicGeo, a training-free framework for high-quality geometric diagram generation. Integrating LLMs with formal solvers for diagram generation, MagicGeo achieves both generalizability and correctness.

\item We introduce a test benchmark to foster research in this area. Empirically, MagicGeo delivers highly accurate geometric diagrams, surpassing the performance of the baseline models, without requiring training data.

\end{itemize}

\begin{figure*}[t]
    \centering
    \includegraphics[width=1.0\textwidth]{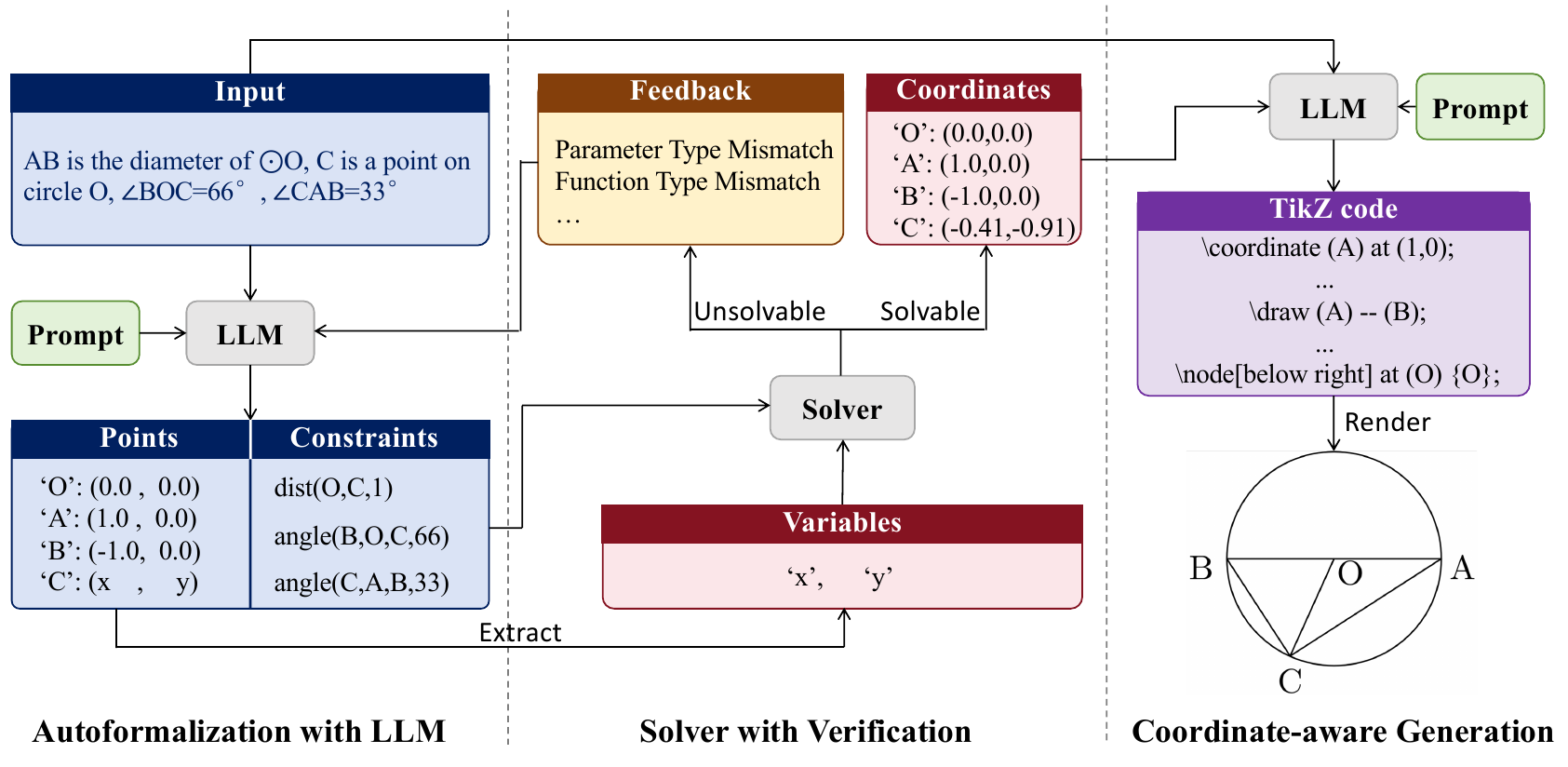} 
    \caption{The overall framework of MagicGeo consists of three stages: Autoformalization with LLM, Solver with Verification, and Coordinate-aware Generation. }
    \vskip -0.1in
    \label{fig:framework}
\end{figure*}

\section{Related Work}


\noindent 
\textbf{Text-to-Image Generation.}
Text-to-image generation~\cite{zhang2023text, bie2024renaissance, jia2024human} has become a rapidly growing field in computer vision and machine learning.
This progress traced back to the emergence of Generative Adversarial Networks (GANs)~\cite{goodfellow2020generative}, which paved the way for research focused on generating images from textual prompts~\cite{reed2016generative, tao2022df, xu2018attngan,zhang2021cross,zhang2017stackgan,zhang2018stackgan++}.
Transformer-based autoregressive models~\cite{ding2021cogview, gafni2022make, ramesh2021zero, yu2022scaling} have attracted significant attention due to their strong capabilities in modeling text-image alignment, as demonstrated by typical models such as DALL-E~\cite{ramesh2021zero} and STAR~\cite{ma2024star}. 
In parallel, diffusion models~\cite{gu2022vector, nichol2021glide,ramesh2022hierarchical, rombach2022high,saharia2022photorealistic} have emerged as a prominent type of generative model for image generation, achieved through the gradual introduction of noise in iterative steps. Notable examples include Imagen~\cite{saharia2022photorealistic}
and others focus on improving compositionality,
e.g., attribute binding~\cite{chefer2023attend,feng2022training}. 

Although these approaches have advanced the generation of realistic scene imagery and propelled text-to-image generation into the spotlight of machine learning research, they struggle with tasks that demand precise control over complex structures and intricate relationships. This includes the generation of diagrams in fields like geometry, architecture, or other technical domains.

\noindent 
\textbf{Text-to-Diagram Generation.}
Generating diagrams from text has long been an intriguing area of research and has recently garnered considerable attention, driven by the success of text-to-image generation. Early efforts~\cite{ghosh2018automated,shahbaz2011automatic, btoush2015generating} primarily focused on generating entity-relationship diagrams, utilizing semantic heuristics to identify entities, attributes, and relationships from natural language specifications. With the rise of LLMs in various language generation tasks~\cite{touvron2023llama, touvron2023llama1, openai2024gpt,chung2024scaling, mann2020language, chowdhery2023palm}, recent work has also leveraged LLMs to facilitate spatial control in diagram generation. 
These methods can be generally classified into two categories: layout-guided models and code-guided methods. Layout-guided approaches, exemplified by DiagrammerGPT~\cite{zala2023diagrammergpt}, employ a two-stage framework that first leverages LLMs to plan layout, then applies layout-guided diffusion models.
Code-guided methods, such as AutomaTikZ~\cite{belouadi2023automatikz}, fine-tune LLMs on large TikZ datasets to generate code for scientific vector graphics, while DiagramAgent~\cite{wei2024words} introduces a four-agent framework leveraging code for text-to-diagram generation and editing.

In geometric diagram generation, both existing approaches face significant limitations. First, image generators suffer from limited spatial fidelity~\cite{gokhale2022benchmarking, chatterjee2024revision, chatterjee2024getting}, despite extensive research in the layout-to-image field~\cite{li2023gligen, yang2023reco,balaji2022ediff, singh2023high, couairon2023zero, xie2023boxdiff}. This limitation prevents these methods from fulfilling precise geometric constraints.
Second, code-guided models for diagram generation are restricted by the capabilities of text-to-code models~\cite{roziere2023code,fried2022incoder,li2022competition,hui2024qwen2,guo2024deepseek}, which rely on large, data-intensive datasets for effective performance.

In contrast, we propose a training-free method that avoids the need for supervised data, leveraging precise point coordinates to enforce stringent geometric constraints. Our approach shares similarities with~\citet{zhengyu2023precise}, which also utilizes point coordinates, but diverges in three key aspects. 1) We leverage the zero-shot capabilities of LLMs to extract points and constraints, bypassing the labor-intensive process of building entity relationship extractors. 2) We introduce a self-verification module to correct LLM-extracted information when the optimization problem is unsolvable. 3) We leverage text-to-code LLMs for TikZ code generation, enabling richer textual insights such as point connections, a capability not fully explored in~\citet{zhengyu2023precise}. Finally, empirical results demonstrate our system's ability of generating complex geometric diagrams.



\section{Method}

In the task of text-to-geometric diagram generation, given a textual description $T$, the objective is to generate a corresponding geometric diagram 
$D$ that adheres to the geometric constraints outlined in $T$.
To realize this objective, we introduce MagicGeo, as depicted in Figure~\ref{fig:framework}. 



\subsection{Autoformalization with LLM}
\label{sec:auto}

We observe that a geometric diagram can be efficiently represented by the coordinates of points and the relationships between them. To formalize this process, we propose a specialized formal language that encapsulates the geometric structure through a set of points and associated constraints, defining their interrelationships. The objective of autoformalization is to convert natural language input, often ambiguous or imprecise, into a precise, unambiguous formal representation that accurately captures geometric relationships and configurations.

Building on the success that LLMs can translate between formal and informal mathematical statements to some extent~\cite{wu2022autoformalization}, we investigate their potential to convert natural language mathematics into our customized formal language, suitable for the solver we introduce. By providing these models with a predefined prompt, we guide their generation, ensuring the output aligns with the requirements of the subsequent solver.

Specifically, we prompt the LLM to generate two key pieces of information: coordinates $Points$ represented by variables $Vars$ and the required geometric constraints $Cons$ based on these coordinates $Points$. 
Figure~\ref{fig:framework} shown an example of autoformalization with LLM.
The prompt consists of a structured database containing a wide range of geometric constraints, along with corresponding instructions that elucidate their precise meanings. By leveraging this structured representation, the LLM interprets the prompt as a comprehensive reference manual, and processes user input in accordance with the specifications outlined in the manual, systematically translating the given descriptions into customized formal languages.

Surprisingly, we find that LLMs exhibit a decent proficiency in formalizing mathematical concepts in our scenario.
Notably, the LLM demonstrates the ability to employ intricate reasoning to adapt and generalize beyond explicitly stated rules. This capability allows the model to infer implicit relationships and make logical extensions where necessary.
For instance, if the input contains the phrase "triangle ABC is inscribed in circle O", the LLM recognizes that this implies the distances from O to points A, B, and C are equal to the radius of the circle. This inference is made despite the absence of explicit instructions in the manual, highlighting the model’s capacity to apply intuitive geometric principles autonomously.


Furthermore, in our approach, we utilize the second phase, namely the solver, to rigorously verify the accuracy of the generated translation. In instances where the candidate autoformalization fails to produce a valid solution, we incorporate the feedback derived from this failure into the process. Specifically, this feedback is treated as a new contextual input, which is then fed into the subsequent iterations of the generation process. This iterative refinement mechanism enables continuous improvement of the formalization output.
Our results demonstrate that by including such a verification step within the framework, the autoformalization accuracy of LLMs is significantly enhanced. 





\subsection{Solver with Verification}
\label{sec:solver}

\noindent 
\textbf{Solver.}
We recognize the existence of numerous interactive theorem provers, such as Isabelle~\cite{wenzel2008isabelle}, Coq~\cite{huet1997coq}, HOL Light~\cite{harrison1996hol, srivas1996formal}, and Lean~\cite{de2015lean, felty2015automated}. These systems function as specialized programming languages, allowing users to formalize statements and construct proofs, which are then automatically verified for correctness.
However, these tools are inherently tailored for mathematical proof problems and thus ill-suited for numerical computation tasks. Additionally, when the solver fails, debugging is challenging due to its lack of interpretability, making it ineffective in guiding the conversational autoformalization process.

To address this, we develop a custom solver, utilizing the constraints of the formal language as function names and leveraging computational analytical geometry methods to examine the constraints and solve the coordinates. Specifically, in order to determine point coordinates, we first identify the relevant variables and extract them into a structured list. We then implement an iterative approach to traverse each variable, simultaneously validating geometric constraints through the derived function names. A precise solution for the coordinates is obtained once a value set is identified for the variables that satisfies all the constraints.

\noindent 
\textbf{Verification.}
Verification plays a crucial role in bridging the Solver and Autoformalization processes, enabling the provision of immediate and actionable feedback for newly generated formalizations. By offering insights into the nature of errors, verification empowers LLMs to refine their understanding and improve the quality of subsequent formalization outputs.
Our experimental analysis highlights two primary failure modes that often require the autoformalization phase to be restarted: (1) detection of non-compliant characters, where symbols or elements violate established syntax or formal language rules, and (2) errors in parameter specifications, including incorrect value assignments or misalignment of parameter numbers.

\begin{figure}[t]
    \centering
    \includegraphics[width=0.48\textwidth]{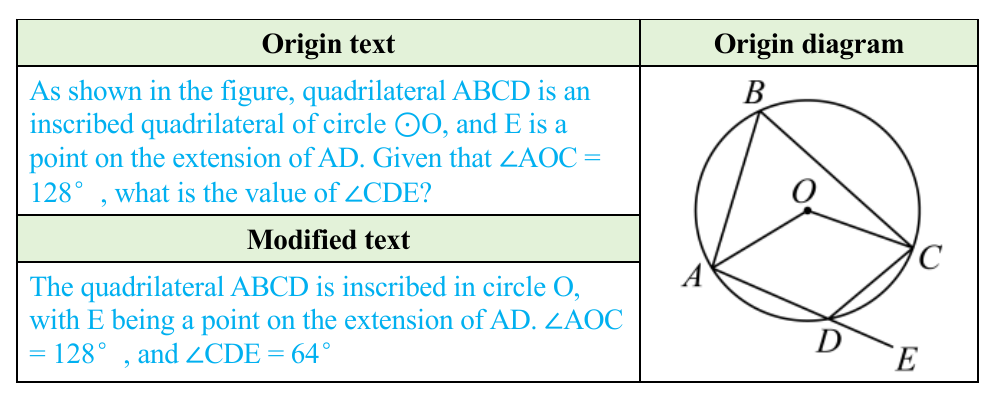} 
    \caption{Illustrating an example of modifying the original text to include necessary information during MagicGeoBench construction. }
    \vskip -0.2in
    \label{fig:modifydata}
\end{figure}

\subsection{Coordinate-aware Generation}
\label{sec:generate}
While directly inputting precise coordinates and textual descriptions into generative models may seem intuitive, it often leads to disorganized visual elements (e.g., misaligned points and lines) that fail to faithfully represent the intended structure. To overcome this limitation, we introduce a more disciplined approach, employing TikZ as an intermediate representation, similar to AutomaTikZ~\cite{belouadi2023automatikz}. However differently, we capitalize on precise point coordinates to harness the zero-shot code generation capabilities of LLMs, eliminating the need for finetuning. This enables the generation of figures that not only maintain structural clarity but also exhibit high fidelity to the original textual descriptions.

\begin{figure*}[t]
    \centering
    \includegraphics[width=1.0\textwidth]{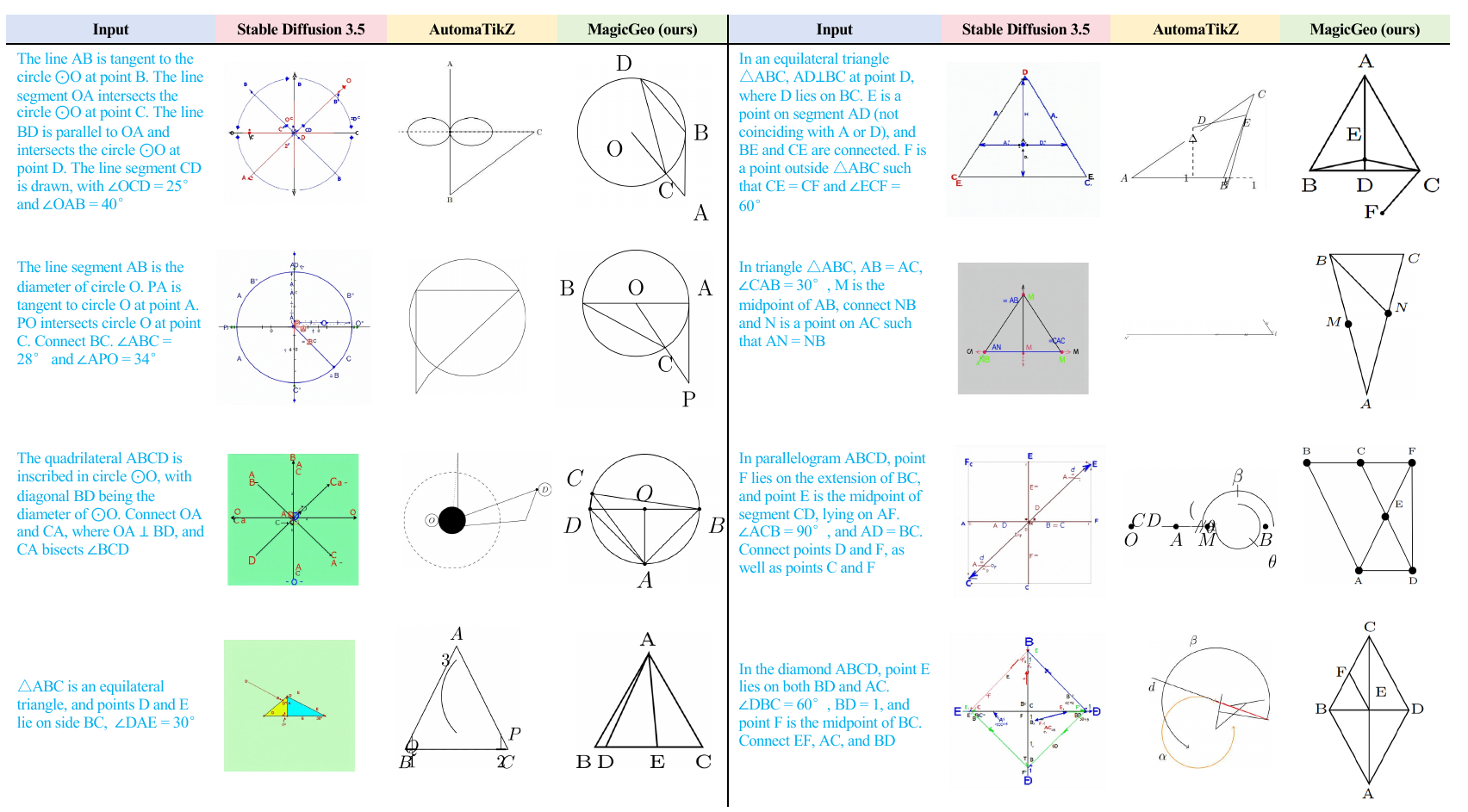} 
    \caption{Qualitative comparison with other approaches. Our method generates results that rigorously adhere to geometric constraints while maintaining high perceptual quality.}
    \label{fig:qualitative}
\end{figure*}

\begin{table*}[ht]
\centering
\small  
\setlength{\tabcolsep}{4pt}  
\renewcommand{\arraystretch}{1.3}  
\begin{tabular}{c |c c c c |c c c c}
\hline
{\multirow{2}{*}{\textbf{ Method}}} & \multicolumn{4}{c|}{\textbf{Img-Txt}}  & \multicolumn{4}{c}{\textbf{Img-Img}} \\
                                 & \textbf{Circle} & \textbf{Triangle} & \textbf{Quadrangle} & \textbf{Avg} & \textbf{Circle} & \textbf{Triangle} & \textbf{Quadrangle} & \textbf{Avg} \\
\hline
SD3.5-Large~\cite{sauer2024fast} & 33.25 & \textbf{34.72} & \textbf{34.02} & \textbf{33.99} & 80.85 & 83.65 & 82.01 & 82.17 \\
AutomaTikZ~\cite{belouadi2023automatikz}          & 30.89 & 29.95 & 29.97 & 30.27 & 84.66 & 83.82 & 87.61 & 85.36 \\
MagicGeo (ours) & \textbf{33.93} & 31.89 & 32.13 & 32.65 & \textbf{91.49} & \textbf{88.16} & \textbf{89.90} & \textbf{89.85} \\
\hline
\end{tabular}
\caption{Quantitative comparison in terms of CLIP score. Higher values indicate better performance (↑).  }
\label{tab:clipscore}
\end{table*}

\section{Experiments}

\subsection{MagicGeoBench}
\label{sec:benchmark}

To rigorously evaluate the performance of text-to-geometric diagram models, we introduce the MagicGeoBench Dataset, a meticulously curated collection of 220 plane geometry questions drawn from high school entrance examinations. In constructing this dataset, we retain the original text for self-contained questions. For questions where essential information is embedded in diagrams rather than explicitly stated in text, we augment the textual descriptions so that diagram can be generated solely from textual input. Figure~\ref{fig:modifydata} illustrates such an example.
The evaluation dataset covers fundamental geometric shapes, and is systematically categorized into three groups: 70 questions on circles, 70 on triangles, and 80 on quadrangles.

\begin{table*}[t]
\centering
\scalebox{0.8}{%
\begin{tabular}{c|cccc|cc c c}
\hline

\multirow{2}{*}{\textbf{Method}} & \multicolumn{4}{c|}{ \textbf{Textual alignment}} & \multicolumn{4}{c}{ \textbf{Image quality}} \\ 
                 & \textbf{Circle} & \textbf{Triangle} & \textbf{Quadrangle} & \textbf{ Avg} & \textbf{Circle} & \textbf{Triangle} & \textbf{Quadrangle}  & \textbf{ Avg} \\
\hline
SD3.5-Large~\cite{sauer2024fast} & 2.50 & 2.20 & 2.72 & 2.47 & 2.42  & 2.11 & 2.11 & 2.21  \\
AutomaTikZ~\cite{belouadi2023automatikz} & 2.28 & 2.08 & 2.12 & 2.16 & 2.28  & 2.08 & 2.15 & 2.17  \\
 MagicGeo (ours) & \textbf{1.22} & \textbf{1.12} & \textbf{1.06} & \textbf{1.13} & \textbf{1.30} & \textbf{1.20} & \textbf{1.20} & \textbf{1.23}  \\
\hline
\end{tabular}%
}
\caption{Average user ranking score of textual alignment and image quality. 1 is the best, 3 is the worst. It is evident that users prefer our results
more than others given the superior quality of ours.}
\label{tab:humanevaluation}
\end{table*}

\subsection{Settings}




\noindent
\textbf{Baselines.}
There is a lack of extensive research focusing on the automatic generation of geometric diagrams. Hence we compare our proposed approach with two established baselines. The first baseline, Stable Diffusion 3.5 (SD3.5)~\cite{sauer2024fast}, is a robust model known for its prowess in generating photorealistic images across a variety of domains. Its ability to synthesize high-quality, realistic images positions it as a strong competitor in the image generation space. The second baseline, AutomaTikZ~\cite{belouadi2023automatikz}, utilizes the TikZ language as an intermediate step for creating high-quality graphical representations, which is a relevant benchmark for our work in the domain of geometric diagram generation. 

\noindent
\textbf{Evaluation Metrics.}
Our goal is to ensure that the diagrams both adhere to textual instructions and conform to typical visual characteristics of geometric illustrations, distinguishing them from photorealistic images. 
To evaluate these two aspects, we utilize CLIP~\cite{radford2021learning} to calculate cosine similarity.
Given that CLIP is designed for general images, we also conduct a user study to validate the effectiveness of our model.

\begin{table}[t]
\centering
\scalebox{0.8}{%
\begin{tabular}{c|cccc}
\hline

                 & {Circle} & {Triangle} & {Quadrangle} & {Avg}\\
\hline
w/o Verification & 92.0 & 95.7 & 96.3 & 94.7 \\
w Verification & \textbf{97.3}& \textbf{100} & \textbf{98.8} & \textbf{98.7} \\
\hline
\end{tabular}%
}
\caption{Illustrating the pivotal role of the
verification mechanism in enhancing the autoformalization process.}
\vskip -0.2in
\label{tab:verification}
\end{table}

\begin{figure}[t]
    \centering
    \includegraphics[width=0.48\textwidth]{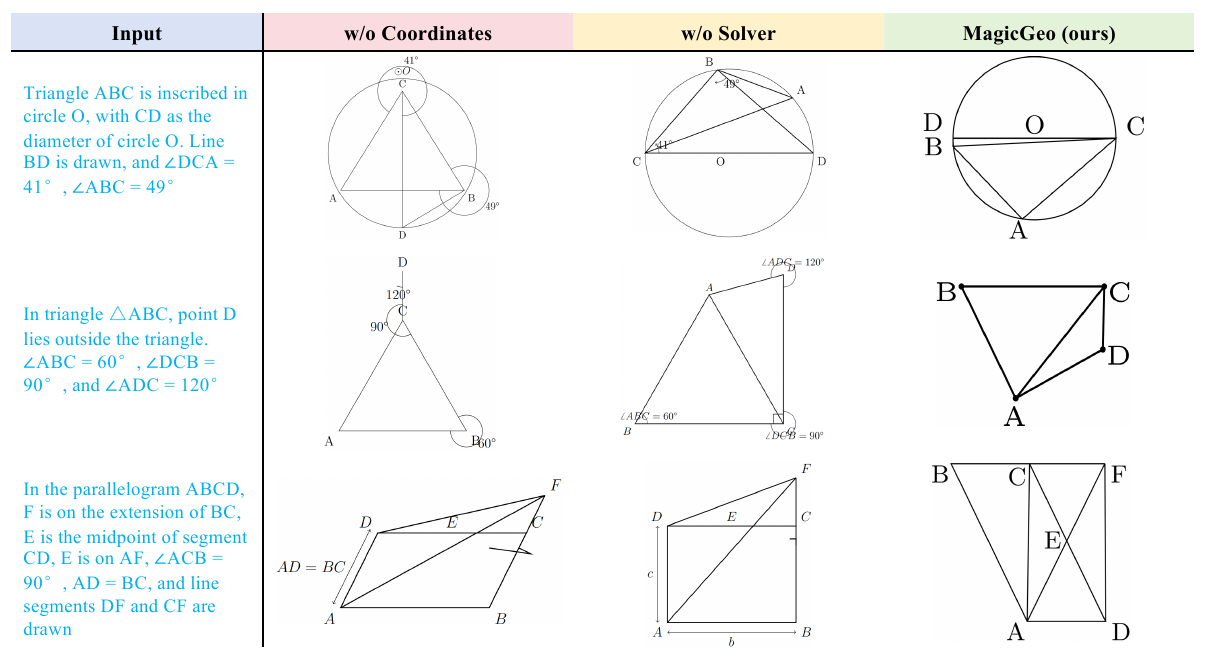} 
    \caption{Illustrating that the solver effectively ensures precise alignment with the accompanying text.}
    \vskip -0.1in
    \label{fig:solver}
\end{figure}

\subsection{Results}

\noindent
\textbf{Quantitative Evaluation.}
Table~\ref{tab:clipscore} presents a comparison between our approach and competitive baselines in terms of CLIP score. When compared to the dedicated image model SD3.5~\cite{sauer2024fast}, our model exhibits a higher similarity to the reference image, as suggested by the image-to-image score. 
However, we notice that in terms of image-text alignment, our model achieves a lower CLIP score. This discrepancy is likely due to the CLIP model’s broad training on general image-text pairs, which may bias towards general image generation models. More importantly, our model surpasses the text-to-diagram baseline AutomaTikZ~\cite{belouadi2023automatikz} in both image-text alignment and image quality, demonstrating the effectiveness of our approach.


\noindent
\textbf{Qualitative Evaluation.}
In addition to quantitative metrics, qualitative evaluation plays a crucial role in assessing generation task. 
We provide the qualitative visual comparison of these methods in Figure~\ref{fig:qualitative}. 
The baseline method SD3.5~\cite{sauer2024fast} generally succeeds in generating simple geometric shapes such as circles and triangles, but it struggles to accurately produce more complex geometric configurations, such as a triangle inscribed within a circle. On the other hand, AutomaTikZ~\cite{belouadi2023automatikz} is capable of generating visually appealing diagrams owing to its use of the TikZ language.
However, both methods fail to consistently adhere to underlying geometric constraints, resulting in diagrams that exhibit noticeable inconsistencies upon inspection.
In contrast, our proposed method rigorously adheres to geometric constraints while simultaneously maintaining a high level of perceptual quality.

\noindent
\textbf{User Study.}
In order to obtain the user’s subjective evaluation of the generated image, we conduct a user study involving 20
participants. In the study, we use 60 samples with 20 in each category. Each sample is consisted of a text input paired with three corresponding output images. 
Participants were instructed to independently rank each image (1 is the best, 3 is the worst) on two distinct aspects: (i) image quality and (ii) adherence to the textual description. 
We report the average
ranking score in Table~\ref{tab:humanevaluation}. It is evident that users prefer our results
more than others given the superior quality of ours.

\begin{table*}[t]
\centering
\setlength{\tabcolsep}{4pt}  
\renewcommand{\arraystretch}{1.3}  
\begin{tabular}{c|c c c c |c c c c}
\hline
{\multirow{2}{*}{{ Method}}} & \multicolumn{4}{c|}{{Img-Txt}}  & \multicolumn{4}{c}{{Img-Img}}  \\
                                 & {Circle} & {Triangle} & {Quadrangle} & {Avg} & {Circle} & {Triangle} & {Quadrangle} & {Avg} \\
\hline
w/o Coordinates            & 31.25 & 31.31 & 30.44 & 31.00 & 86.31 & 88.46 & 88.90 & 87.89 \\
w/o Solver            & 32.25 & 31.52 & 30.77 & 31.51 & 87.37 & \textbf{88.51}  & \textbf{90.84}  & 88.91 \\
MagicGeo (ours) & \textbf{33.93} & \textbf{31.89}  & \textbf{32.13}  & \textbf{32.65}  & \textbf{91.49}  & 88.16 & 89.90 & \textbf{89.85}  \\
\hline
\end{tabular}
\caption{The effect of solver in terms of CLIP score. Using LLM to directly generate TikZ code is denoted as w/o Coordinates. Asking LLM to infer coordinates followed by coordinate-aware generation is denoted as w/o Solver.}
\vskip -0.1in
\label{tab:ablation_clipscore}
\end{table*}

\subsection{Ablations}

\noindent
\textbf{The Effect of Verification.}
In our framework, when the solver fails to find a solution, feedback is provided to the LLM for re-autoformalization, a process we refer to as verification. This process allows for a maximum of five feedback iterations, aiming to iteratively correct errors identified by the solver. Here we compare its performance against a baseline system that excludes verification. The evaluation criterion focuses on the accuracy of output points and constraints, which are manually verified. As shown in Table~\ref{tab:verification}, the incorporation of verification results in a substantial improvement in autoformalization accuracy, increasing from 94.7\% to 98.7\%. This highlights the pivotal role of the solver’s feedback mechanism in enhancing the autoformalization process.

\noindent
\textbf{The Effect of Solver.}
We propose the use of analytical geometry methods to develop a custom solver designed for precise point location determination, which is subsequently leveraged for diagram generation. To evaluate the effectiveness of our solver, we compare our approach against two alternative methods:
(1) w/o Coordinates: this approach utilize LLMs to directly generate TikZ code without incorporating explicit point coordinates, akin to the approach used in AutomaTikZ;
(2) w/o Solver: this variant first ask the LLM to infer the point coordinates and then use these coordinates for coordinate-aware TikZ generation.
We compare their results in terms of the CLIP score in Table~\ref{tab:ablation_clipscore}. 
To provide a clearer understanding, we present several visual examples in Figure~\ref{fig:solver}.
The incorporation of explicit coordinates significantly enhances the quality of diagram generation. However, some issues persist, such as the failure to satisfy the constraint "angle ABC equals 49 degrees" in the first example and the constraint "angle ADC equals 120 degrees" in the second example. In contrast, the application of solver effectively addresses all constraints, ensuring precise alignment with the accompanying text and thus superior quality. 

\begin{table}[t]
\centering
\scalebox{0.8}{
\renewcommand{\arraystretch}{1.3}  
\begin{tabular}{c|c|cccc}
\hline
{{Stage}} & {{LLM}} & \multicolumn{1}{c}{{Circle}} & \multicolumn{1}{c}{{Triangle}} & \multicolumn{1}{c}{{Quad}} & \multicolumn{1}{c}{{Avg}}\\ 
\hline
\multirow{3}{*}{1}& DeepSeek-V3  &  {0.97} &  {1.00} &  {0.99} &  {0.99} \\
&Qwen-plus & 0.96  & 0.99 & 0.99 & 0.98  \\
&GPT-4o mini       & {0.97} & 0.99  & 0.99 & 0.98\\
\hline
\multirow{3}{*}{3}& DeepSeek-V3 &  {1.00}  &  {1.00}  &  {1.00}  &  {1.00}  \\
&Qwen-plus & 1.00  & 1.00 & 0.97 &  0.99 \\
&GPT-4o mini      & 0.99 & 1.00  & 1.00 & 1.00\\
\hline
\end{tabular}%
}
\caption{Illustrating that our framework is robust to different LLMs, which shows negligible impact.}
\vskip -0.2in
\label{tab:LLMs}
\end{table}

\noindent
\textbf{The Effect of Using Different LLMs.}
We employ the DeepSeek-V3~\cite{liu2024deepseek} model for both Stage 1 and Stage 3 in our framework. To study the impact of different LLMs, we investigate two models: Qwen-plus~\cite{yang2024qwen2} and GPT-4o mini~\cite{shahriar2024putting} . We isolate the LLM variation to a single stage—either Stage 1 or Stage 3—while maintaining the other stage constant.
For evaluation, we manually examine autoformalization accuracy in Stage 1 and visually inspect the generated diagrams in Stage 3. A sample of 60 instances was experimented, with the accuracy presented in Table~\ref{tab:LLMs}.
It is important to note that we utilize distinct prompts for different LLMs to fully harness their respective capabilities.
Our findings indicate that the choice of LLM has a negligible impact on the final outcomes, demonstrating the suitability of LLMs for these tasks. This suggests that our framework maintains consistent performance regardless of the specific LLM.

\begin{figure}[t]
    \centering
    \includegraphics[width=0.49\textwidth]{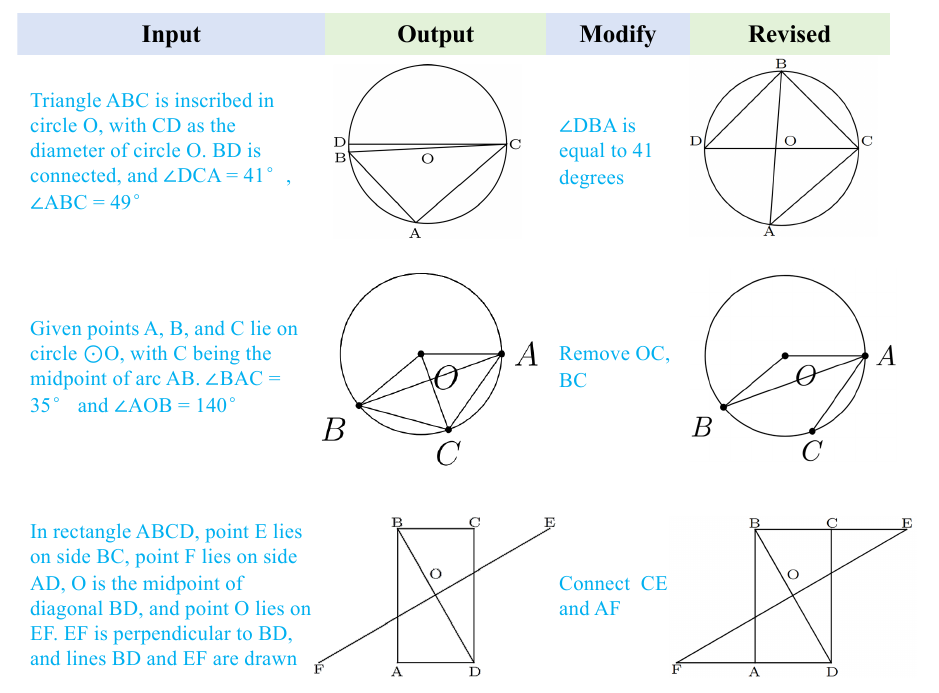} 
    \caption{Application to diagram editing. }
    \vskip -0.1in
    \label{fig:application}
\end{figure}

\subsection{Application to Diagram Editing}
Our method leverages precise coordinate information and thus is able to make effective diagram modifications based on user intent. We present examples of diagram editing results in Figure~\ref{fig:application}. Simple tasks, such as adding or deleting lines, are accomplished by re-executing the third stage. More complex adjustments(e.g., specify a new angle degree in the first example), are handled effectively by our framework, which quickly determines the necessary coordinates for adjustment.
This demonstrates the potential of our framework in real-world diagram editing applications.

\section{Conclusion}

In conclusion, this paper presents MagicGeo, a novel framework for the automatic generation of geometric diagrams from textual descriptions, which stands out for its training-free approach and high precision. By reframing the diagram generation task as an optimization problem, MagicGeo ensures the accuracy of key geometric properties—such as parallelism and orthogonality—by leveraging analytical geometry rules.
The comprehensive evaluation of MagicGeo, including empirical comparisons with state-of-the-art baselines and ablation studies, demonstrates its effectiveness in producing accurate and reliable diagrams. Ultimately, MagicGeo offers significant potential for streamlining the creation of educational and academic diagrams, with broader implications for enhancing content generation in scientific and educational settings.

\section{Limitations}

While MagicGeo demonstrates notable advancements in the automatic generation of geometric diagrams from textual descriptions, several limitations must be acknowledged.

One limitation of our framework is its reliance on LLMs to translate complex geometric descriptions into formal representations that adhere to geometric conventions and generate accurate TikZ code. Although current translation performance, as shown in the ablation, is highly effective, it is not yet flawless, with visual examples presented in the Appendix section. We anticipate that ongoing advancements in LLM research, particularly in mathematical reasoning and code generation, will mitigate this limitation.

The current solver exhibits extended processing times for complex diagrams, with efficiency influenced by factors such as input complexity, the number of geometric entities, and the precision required for diagram generation. Preliminary experiments indicate that generation times typically range in the order of seconds; while for very intricate complex geometric diagram, processing times can exceed one hour. Future work will focus on enhancing solver efficiency through parallelization, optimized constraint-solving methods, and the development of heuristic techniques that balance computational cost and diagram accuracy.










\bibliography{custom}

\end{document}